\documentclass[10pt,twocolumn,letterpaper]{article}

\usepackage{wacv}
\usepackage{times}
\usepackage{epsfig}
\usepackage{graphicx}
\usepackage{amsmath}
\usepackage{amssymb}

\usepackage[english]{babel}
\usepackage{amsmath}
\usepackage{amssymb}
\usepackage{calc}
\usepackage[caption=false]{subfig}
\usepackage{cite}

\usepackage{makecell}
\usepackage{fancyhdr}

\DeclareMathOperator*{\argmax}{arg\,max} 

\graphicspath{{figures/}}

\wacvfinalcopy 

\def\wacvPaperID{412} 

\ifwacvfinal\fi
\begin{document}
\fancypagestyle{plain}{
\fancyhf{} 
\fancyfoot[L]{Copyright IEEE 2018}
\fancyfoot[C]{}
\fancyfoot[R]{}
\fancyhead[C]{To appear as a conference paper at the 2018 IEEE Winter Conf. on Applications of Computer Vision (WACV’2018)}
\renewcommand{\headrulewidth}{0pt}
\renewcommand{\footrulewidth}{0pt}
}  

\title{Activity-conditioned continuous human pose estimation for performance analysis of athletes using the example of swimming}

\author{Moritz Einfalt \hspace{2cm} Dan Zecha \hspace{2cm} Rainer Lienhart\\
Multimedia Computing and Computer Vision Lab, University of Augsburg\\
{\tt\small moritz.einfalt@informatik.uni-augsburg.de}
}

\maketitle

\begin{abstract}
In this paper we consider the problem of human pose estimation in real-world videos 
of swimmers. Swimming channels allow filming swimmers simultaneously 
above and below the water surface with a single stationary camera.
These recordings can be used to quantitatively assess the athletes' performance. The quantitative evaluation, so far, 
requires manual annotations of body parts in each video frame.
We therefore apply the concept of CNNs in order to automatically infer the required pose information.
Starting with an off-the-shelf architecture, we develop extensions to leverage activity information 
-- in our case the swimming style of an athlete -- and the continuous nature of the video recordings.
Our main contributions are threefold: 
(a) We apply and evaluate a fine-tuned Convolutional Pose Machine architecture as a baseline in our very challenging aquatic environment and discuss its error modes, 
(b) we propose an extension to input swimming style information into the fully convolutional architecture and 
(c) modify the architecture for continuous
pose estimation in videos. 
With these additions we achieve reliable pose estimates with up to $+16\%$ more
correct body joint detections compared to the baseline architecture.
\end{abstract}

\section{Introduction}
\label{intro}

\begin{figure}[t]
  \noindent \begin{centering}
    \includegraphics[width=0.47\textwidth, keepaspectratio]{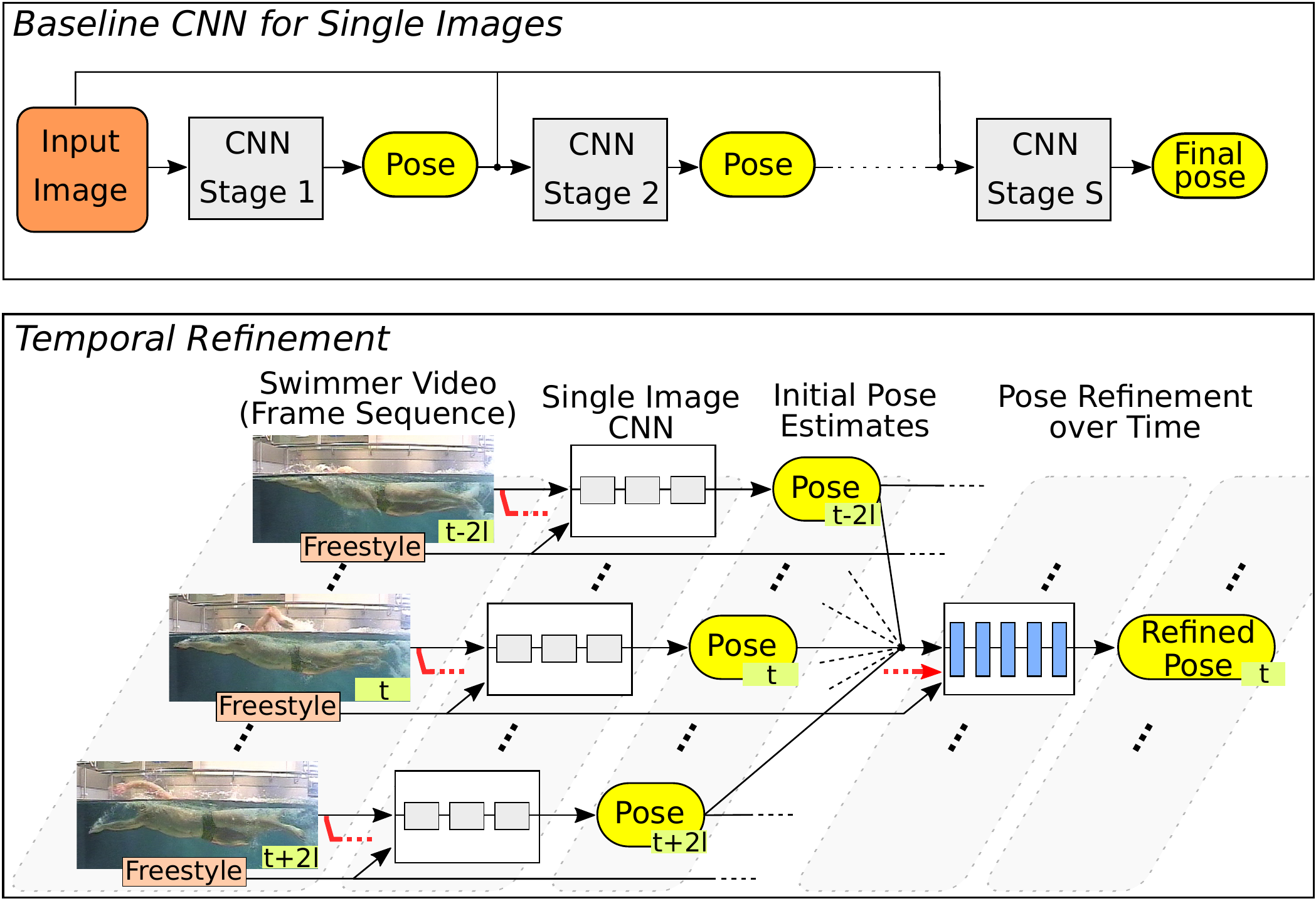}
    \caption[Overview]{
      \label{fig:overview} Overview of our approach to reliable human pose estimation in swimming channel videos. Top: Baseline CNN from \cite{Wei2016_CPM} for human pose estimation on single images. Bottom: Our architecture uses initial estimates from the baseline CNN and refines these over fixed length pose sequences. Network layers are conditioned on the swimming style of an athlete.}

    \par\end{centering}
\end{figure}

In recent years, an increasing interest in computer vision applications
in the sports domain can be observed. One important reason is that broadcasts
of sport events are among the most popular content on TV and the
internet \cite{Turchini2015_Sport_Activities}. The footage offers
plenty of possibilities to gather additional sta\-tis\-tics, including team performance
statistics for the participating athletes and their coaches. 
In this work we focus on the specific scenario of video recordings of top-tier 
swimming athletes in a swimming chan\-nel.
In competitive swimming, such swimming channels can be used for individual
performance analysis. They consist of a pool with
an adjustable artificial water current, flowing in a fixed direction.
Cameras are positioned at various locations around the pool, both
above and below the water surface, to record the athlete in the channel.
By matching the flow velocity, the athlete can perform normal swimming
motion while staying in the same position relative to the cameras.
This enables the recording of swimming motion over a long period of time.
The recordings are then used by an expert of the field to quantitatively asses
 the performance
and to work out possible improvements of the athlete's technique and further training strategies. Inferring
information about the athlete's movements such as the stroke rate or other
intra-cyclic kinematic parameters over time requires to annotate the video material appropriately.
These annotations can range from a sparse selection of frames showing
characteristic key-poses to frame-by-frame locations
of the athletes body parts and/or joints \cite{Zecha2015_KeyPose}. 
Automating this process by employing a vision-based pose 
estimation system can alleviate the massive overhead
of manual pose annotations. Moreover, it opens up new kinds of individual 
training support to even more athletes as much more regular quantitative data
is available for performance and improvement analysis.

Human pose estimation is a long standing task in the domain
of computer vision.
There exists a large research community that is constantly striving
to develop better and more reliable methods. With the continuing success of Convolutional Neural Networks (CNNs), 
several architectures for end-to-end human pose estimation have emerged, 
including the Convolutional Pose Machines (CPMs) of Wei et al. \cite{Wei2016_CPM}.
Based on their notable success and influence on other architectures \cite{Newell2016_Hourglass,Insafutdinov2016_DeeperCut}, 
we apply the CPM framework to our 
challenging scenario of human pose estimation in swimming channel 
recordings and analyze its performance and failure modes.
Based on these, we propose architectural improvements to (a) leverage 
additional activity information including the swimming style of an athlete
and to (b) exploit the temporal connection of poses in consecutive frames,
i.e., the fact that poses are estimated in temporally and causally related video frames. Figure~\ref{fig:overview} gives an overview of our approach.

\section{Related work}
\label{sec:rel_work}

Computer vision has been adopted for various applications in the sports domain.
Prominent tasks include sports type \cite{Gade2013_SportsType} and activity
recognition \cite{Turchini2015_Sport_Activities,Nibali_2017_CVPR_Workshops}, tracking athletes
and other objects of interest in videos \cite{Soomro_2015_Tracking,Wei2015_Ball_Ownership}
and human pose estimation \cite{Fastovets2013_PE,Hwang_2017_CVPR_Workshops}. \cite{Moeslund2015_CV_Sports}
offer an overview of a wide range of application.

For performance analysis of individual athletes, \cite{Hasegawa2016} propose a method
to facilitate speed and stride length estimation for runners based on hand-held camera recordings.
Specific to an aquatic environment, \cite{Sha2013_swimmer_loc} describe how swimmers can be tracked 
when filmed by a moving camera above the water surface.
\cite{Zecha2015_KeyPose} discuss the identification of characteristic poses 
of swimmers.
\cite{Victor_2017_CVPR_Workshops} present a CNN approach for automatic stroke-rate estimation in swimmer videos.

The traditional approach to human pose estimation are pictorial structures, 
where the human body is modeled as a collection of interacting parts 
\cite{Felzenszwalb2005_PictStructs,andriluka2009_PictStructs_Revisited,Johnson10_LSP,Sapp2013_MODEC,Wang_2013_CVPR}.
The model describes the appearance of individual parts and the relationship between interacting parts in a probabilistic fashion.
The goal is to find the most probable part configuration given an input image.

Recent literature focuses on methods using CNNs to overcome the drawbacks of hand-crafted image features and 
limited part interactions present in classical approaches.
The currently best results on popular human pose estimation benchmarks 
like the Leeds Sports Pose (LSP) \cite{Johnson10_LSP}  and MPII Human Pose \cite{Andriluka2014_MPII} datasets all apply CNNs.
\cite{Toshev2014_DeepPose} describe an architecture that directly re\-gres\-ses the image coordinates of body joints.
Subsequent publications regress confidence maps that indicate the likelihood of all possible joint locations in an image 
\cite{Ramakrishna2014_PoseMachines,Pfister2015_Flowing_ConvNets,Newell2016_Hourglass,Hwang_2017_CVPR_Workshops}.
This spatial encoding of the learning objective seems to be more natural to CNNs compared to the direct regression of image coordinates.
Another common design are architectures performing iterative refinement 
\cite{Carreira2016_error_feedback,Belagiannis2016_Recurrent_HPE,Cao2016_CPM_MultiPerson}.
After producing an initial pose estimate it is progressively refined in the deeper layers of the network.
There are also proposals to use classical part-based models to refine the pose estimates from CNN-based methods, 
either as a separate post-processing step \cite{Pishchulin2016_DeepCut} 
or by mapping the domain-specific part interactions into the neural network itself for an end-to-end trainable architecture \cite{Yang_2016_CVPR}.

While most publications focus on human pose estimation on single 2D images, we are additionally interested in human pose estimation on videos.
\cite{Fastovets2013_PE,Zhang_2015_ICCV} use pictorial structures to model humans in videos. 
They extend the spatial interactions between body parts by temporal dependencies that describe the change of body part configurations over time.
Flowing Conv-Nets \cite{Pfister2015_Flowing_ConvNets} combine a CNN for human pose estimation on single images with a second CNN for the optical flow in videos that enables an estimate of the movement of body parts.
In \cite{Song_2017_CVPR}, optical flow and both spatial and temporal part interactions are used jointly in a single network architecture. \cite{Gkioxari_2016} describe a recurrent neural network (RNN) architecture applied to sequential video frames. In our approach we avoid the computational expensive extraction of optical flow and the data-intensive training of RNNs due to limited video material.

\section{Human pose estimation for swimming channel recordings}

\begin{figure}[t]
\noindent \begin{centering}
\subfloat[\label{fig:challenge_noise}Backstroke]{\noindent \begin{centering}
\includegraphics[width=0.22\textwidth, keepaspectratio]{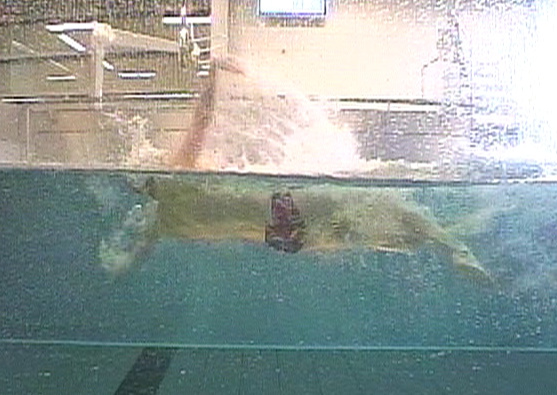}
\par\end{centering}

}\subfloat[\label{fig:challenge_refraction}Breaststroke]{\noindent \begin{centering}
\includegraphics[width=0.22\textwidth, keepaspectratio]{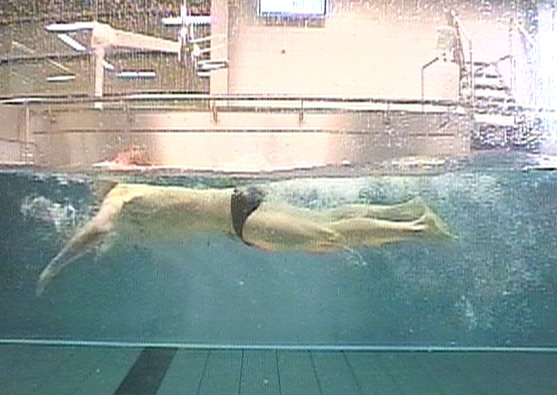}
\par\end{centering}

}
\par\end{centering}

\noindent \begin{centering}
\subfloat[\label{fig:challenge_occlusion}Butterfly]{\noindent \begin{centering}
\includegraphics[width=0.22\textwidth, keepaspectratio]{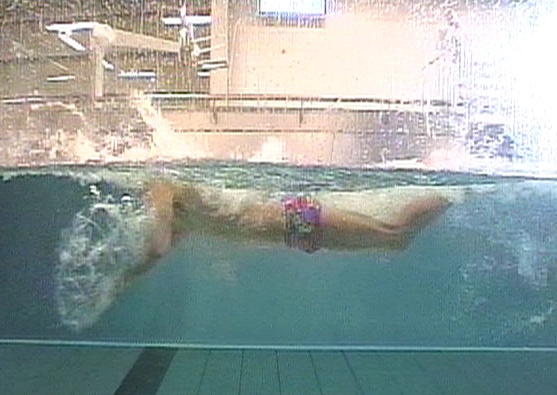}
\par\end{centering}

}\subfloat[\label{fig:challenge_self_occlusion}Freestyle]{\noindent \begin{centering}
\includegraphics[width=0.22\textwidth, keepaspectratio]{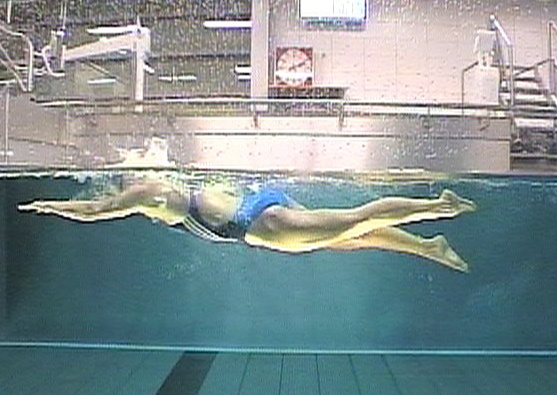}
\par\end{centering}

}
\par\end{centering}

\caption[Exemplary video frames from the swimmer dataset.]{
  \label{fig:swimstyle_examples}Exemplary frames from the swimming
  channel footage. The data poses multiple challenges: (a) Image noise
  due to bubbles and spray. (b) Ambiguous joint locations (head) due
  to refractions at the water surface. (c) Occlusion by the water line
  (head, lower legs). (d) Self-occlusion (left arm).}
  
\end{figure}

The video material used in this work
has been recorded at $720\times576 @ 50$i with a single stationary camera behind a glass pane
at the left side of a swimming channel. 
The athletes are depicted from the side, partially above and below the water surface. The swimming direction is always right to left.
The videos display different male and female athletes at different flow velocities performing
four different swimming styles: backstroke, breaststroke, butterfly
and freestyle. The swimming style throughout each video does not change.
Figure~\ref{fig:swimstyle_examples} shows exemplary video
frames for all four styles. 

The video frames are annotated using a person-centric, $14$ joint body model.
For the symmetric swimming styles such as breaststroke and butterfly only the
left side of the body is annotated by hand. Due to the side-view,
the body parts on the right side are usually directly occluded by their left
counterpart. Using the same image coordinates for both left and right joints
is thus a good approximation in most cases. For backstroke and
freestyle (anti-symmetrical motion) all $14$ joints are annotated
explicitly.

The viewpoint of the camera and the underwater setting present multiple
challenges for human pose estimation:
\begin{itemize}
\item Image noise due to bubbles and spray (Figure~\ref{fig:challenge_noise}).
\item Ambiguous joint locations due to refraction at the water surface 
(Figure~\ref{fig:challenge_refraction}).
\item Occluded joints by the water surface (Figure~\ref{fig:challenge_occlusion}).
\item Frequent self-occlusions (Figure~\ref{fig:challenge_self_occlusion}).
\end{itemize}
This makes precise joint localization difficult, even for a human
observer. At the same time, the variety of poses, appearance and background
is rather limited compared to unrestricted human pose estimation in the wild.

\subsection{Baseline approach}
As a baseline method we use the CPM framework from \cite{Wei2016_CPM} for human pose estimation on single images. 
It uses a pure convolutional neural network divided into $S$ identical
stages. The network is trained on instances $(x,\mathbf{y})$, where
$x$ is the input RGB image of fixed size, centered on the person
of interest. $\mathbf{y}=(y_{1},\dotsc y_{J})$ represents the
ground truth locations of all $J$ joints in Cartesian image coordinates.
The objective of the network is to regress confidence values for all
possible joint locations. The output of every stage $s \in [1,S]$ is a stack
of confidence maps (in the following simply denoted as \emph{heatmaps}) $\hat{\mathbf{h}}^{s}=(\hat{h}_{1}^{s},\dotsc,\hat{h}_{J}^{s})$.
For every image location $z \in Z$, $\hat{h}_{j}^{s}(z)$ is interpreted
as the confidence that joint $j$ is located at $z$. The locations with highest confidence
in the final set of heatmaps $\hat{\mathbf{h}}^{S}$  are used as the predicted locations $\hat{y}_{j}$
for each joint $j$:
\begin{equation}
  \hat{y}_{j} = \argmax_{z\in Z} \hat{h}_{j}^{S}(z).
\end{equation}
The main motivation for using a deep, stage-wise CNN architecture
is to utilize spatial context and learn dependencies 
between any set of joints. This is necessary for improved estimates on difficult
instances, e.g. when some body parts are occluded or when distinguishing left-side body parts from their right-side counterparts. 
The first network stage uses the input image 
to predict a first set of joint heatmaps, which is then subsequently refined by the following stages (see Figure~\ref{fig:overview}). 
The increasing receptive field and thus
the additional spatial context with each subsequent stage enables the network to resolve ambiguities in
the estimates of previous stages. In this work we consider a fine-tuned CPM with three stages, initialized on general human pose estimation in sports. It operates on each video frame individually and acts as the baseline architecture.

\section{Utilizing swimming style information}
\label{sec:utilizing_swimming_style}

A key challenge in our aquatic environment are frequently occluded body parts, either hidden behind 
other parts of the body or the waterline. 
Detecting these body parts is necessary to enable the evaluation of an athletes movement over time, 
\eg for inferring kinematic parameters like angular and vectorial velocities. 
Despite the CNNs ability to implicitly learn joint and body part dependencies, frequent occlusions can make an unambiguous reconstruction of an athletes pose very difficult. 
Figure~\ref{fig:challenge_self_occlusion} shows an exemplary video frame, where the athletes left arm is not directly visible. 
Given only this single frame, even a human observer might not be able to correctly estimate the pose. 
However, the pose of a swimmer is directly related to his swimming style.
 Information about his swimming style can resolve the ambiguous location of the left arm, 
 since \eg the positioning of both arms on top of each other is unlikely for a freestyle swimmer. 
 We therefore want to encode this additional contextual information into the CNN architecture 
 as it further reduces the space of possible poses and enables the learning of more specific body part dependencies. In Section~\ref{sec:seq_pose_refinement} we will also exploit the temporal aspect of motion to improve further.

 \subsection{One-hot class label maps}
 \label{sec:one_hot}
The additional swimming style information partitions the data into four distinct classes. 
Each video and thus each frame
is labeled to belong to one of these classes.
We propose a very simple way to encode class labels by using additional 2D input channels,
one for each class, and call them \emph{class label maps}.
The class label maps are one-hot encoded, \ie, the respective map of a given swimming style is filled with ones, 
while the remaining three maps are zeroed.
These additional maps can be added as supplementary input channels to one
or multiple convolution layers in the network.
We choose this encoding to retain a pure convolutional architecture
in favor of \eg the addition of an encode-decoder network block where
an abstract hidden state can be combined with
arbitrary additional input information \cite{Tatarchenko2016_Multi_View}.

When adding class label maps to a convolutional layer, the new filter weights shift the filter activations 
by a constant but class dependent value (since the class label input is constant as well).
This is identical to an additional class dependent bias term learned for each filter.
With a subsequent ReLU activation function, the filter activations are either increased, decreased or set to zero when falling below the ReLU threshold.
The observed effect is an on-off switching functionality:
Activations of a filter surpass the threshold only if a specific subset of class labels is given.
It enables the CNN to learn general purpose filters as well as swimming style specific ones and to combine them appropriately.

\begin{figure}[t]
  \noindent \begin{centering}
    \includegraphics[width=0.47\textwidth, keepaspectratio]{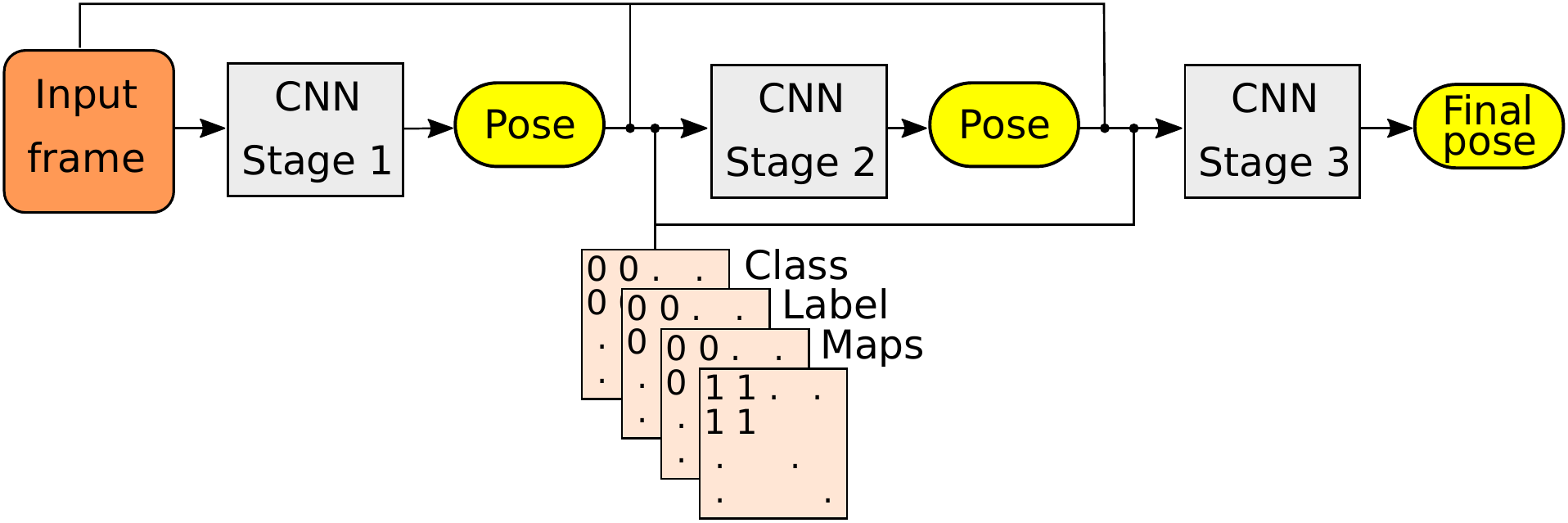}
    \caption[CPM architecture using class label maps.]{
      \label{fig:class_label_architecture} ``Style input - once'' architecture using one-hot encoded class label maps as additional input.}

    \par\end{centering}
\end{figure}

The benefit of swimming style information is that it reduces the variety
of expected poses and enables learning of more powerful
dependencies between joints. Adding the swimming style input at the beginning of the baseline architecture might thus not be useful, 
as the layers of the first network stage operate on local
image content with a limited receptive field. The subsequent stages
seem more appropriate, as they have the capability to learn long-range
spatial dependencies between different joints and can naturally benefit
from the additional input. We evaluate two strategies for adding swimming style
class label maps to the
baseline architecture. The first approach is to add class label maps once at the beginning of each stage $s>2$, as depicted in Figure~\ref{fig:class_label_architecture}. 
The network has to propagate the class information manually if it is required in deeper layers. 
In the second approach, class label maps are added to all subsequent network layers as well, 
effectively replicating the class information. We denote the two variants ``Style input - once'' and ``Style input - repeated''.

\section{Temporal pose refinement}
\label{sec:seq_pose_refinement}
For our baseline approach, we naively perform human pose estimation on single video frames,
ignoring the temporal component of the video data.
The same is true for the swimming style conditioned variant,
where the only information shared across time is the constant swimming style in each video. 
While the pose of an athlete is not constant, the changes from one frame to another
are limited. This is especially true for the swimmer videos recorded
at $50$Hz. Thus, there is a strong dependency between the pose in
the current frame and poses in past and future frames. If past
poses are known, they can act as an important cue for what
pose to expect in the next frame(s).

Given the visual challenges in swimmer videos (see Figure~\ref{fig:swimstyle_examples}), 
precise estimation of
joint locations on single frames can be difficult. Sequential video
frames enable humans to estimate the pose of athletes with higher precision by
interpolating the location of joints (and thus their movement) over time.
We propose an architectural extension to the baseline CNN that follows this idea of
temporal refinement.

\subsection{Temporal sequence model}
The stage-wise architecture of the
baseline CNN is intended to facilitate learning of spatial dependencies.
In the same manner, one can imagine an additional network stage to learn
temporal dependencies. We propose to exploit the temporal dependencies by a stage that uses the estimated
poses on past and future video frames to improve the pose for the
current one. To avoid explicit recurrence and enable pose refinement
both forward and backward in time, this stage is 
a separate CNN operating on sequences of single-frame pose estimates. It acts
as an additional post-processing network that refines pose estimates
over time.

\begin{figure*}[t]
  \centering{}
  \includegraphics{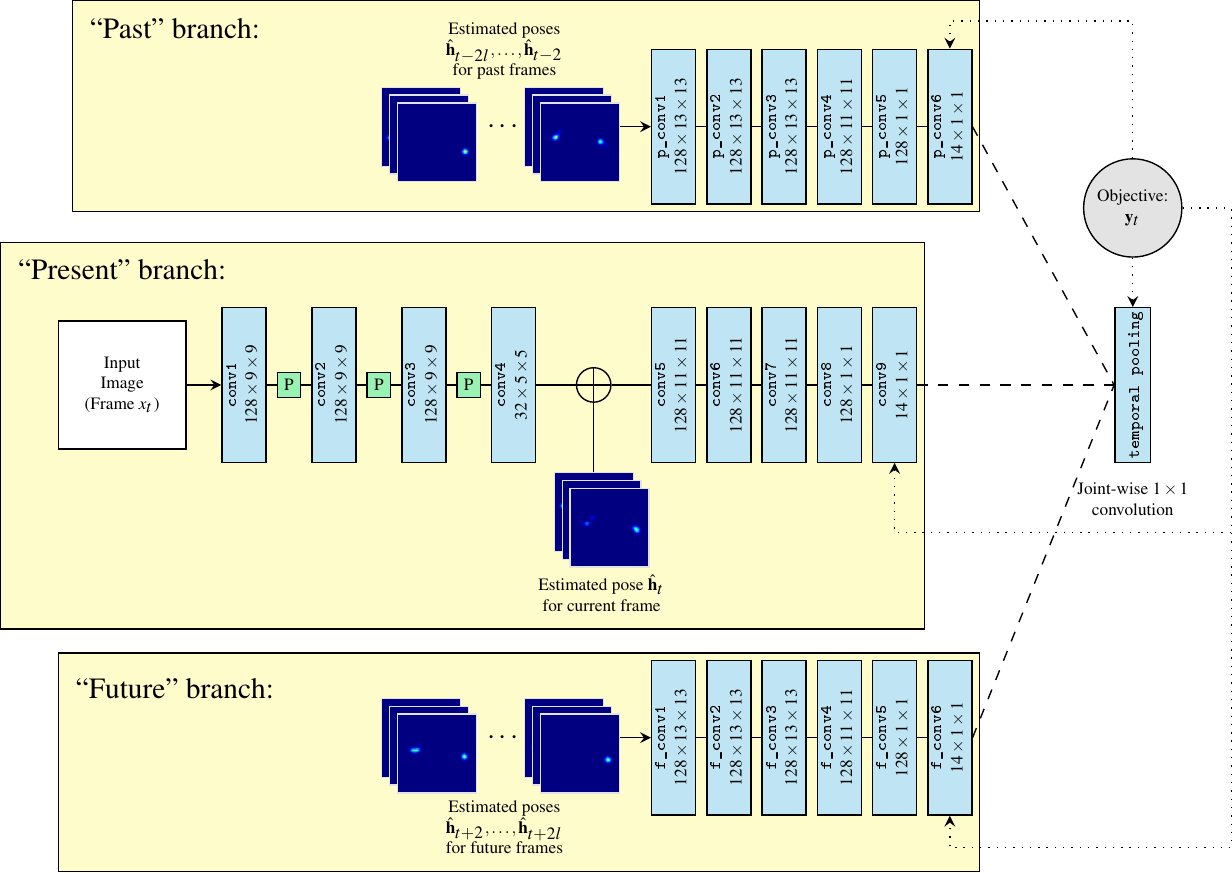}
  \caption[Sequential pose refinement architecture with temporal pooling.]{
    \label{fig:temp_pooling_architecture}Network architecture for temporal
    pose refinement. Pose estimates from past and
    future frames are processed in separate network branches. Each branch
    is trained to predict heatmaps representing the pose $\mathbf{y}_t$ for the current frame $x_t$. The temporal
    pooling layer combines predicted heatmaps from all three branches
    and is trained separately. \textbf{P} denotes $\times 2$ max. pooling.}
\end{figure*}

Our overall architecture for human pose estimation on videos consists
of the baseline 3-stage CNN and the temporal post-processing network
that we now describe.
Each video consists of frames $(x_{1},\dotsc,x_{T})$, where $T\in\mathbb{N}$ is the length the video. We apply the baseline CNN to each frame $x_{t},$ $t\in[1,T]$, and obtain a single-frame
pose estimate in the form of localization heatmaps $\hat{\mathbf{h}}_{t}=(\hat{h}_{t,1},\dotsc,\hat{h}_{t,J})$
for the $J=14$ joints. Our post-processing network
uses frame $x_{t}$ and the \emph{sequence of pose estimates $\mathbf{z}_{t}$
}with 
\begin{equation}
\mathbf{z}_{t} = (\hat{\mathbf{h}}_{t-2l},\hat{\mathbf{h}}_{t-2(l-1)},\dotsc,\hat{\mathbf{h}}_{t+2l})\label{eq:input_sequence}
\end{equation}
as input. It outputs a refined estimate $\mathbf{h}_{t}^{*}$, which
is again a set of $J$ joint localization heatmaps for the frame
at time $t$. The free parameter $l\in\mathbb{N}$
defines how many single-frame estimates from
past and future frames are used to refine a pose.
We denote the length of the sequence by
$k$ with $k=4l+1$.
Note that we use pose sequences with reduced temporal resolution
which contain pose estimates from every other frame only. This leads to an effective input sequence length of $k^{\prime}=2l+1$ and enables long pose sequences for refinement while keeping the input size feasible.

\subsection{Network architecture for temporal interpolation of joint predictions}
\label{sec:temp_refinement_net}

The proposed temporal refinement network depicted in Figure~\ref{fig:temp_pooling_architecture} takes $(x_{t},\mathbf{z}_{t})$ as its input
and is trained to predict the ground truth joint location $\mathbf{y}_{t}$.
However, the input is not simply stacked and processed jointly.
The network architecture is designed to reflect the idea of temporal interpolation of
joint locations based on predicted locations in surrounding video frames,
\ie based only on $\mathbf{z}_{t}$.
We enforce this mechanism by dividing the network input into three parts: the
single-frame pose estimates for past frames, the estimates for future
frames and the estimate for the current frame together with the video
frame itself. All three parts are processed in different network branches.
Each branch is trained to predict the same target \textendash{} the
pose for the current frame. The ``past'' and ``future''
branches do not possess any image information and are thus forced
to predict the current joint locations solely based on the joint estimates
in the preceding or subsequent frames. This requires to infer some
notion of interpolation forward or backward in time.
The ``present'' branch does not have
any temporal information and resembles a normal network stage in the baseline 
architecture. The outer branches use convolution layers
with larger filter kernels for an increased receptive field.

The final layer of the network is intended to integrate the predictions
of all three branches. It is based on the idea of \emph{temporal pooling}
in \cite{Pfister2015_Flowing_ConvNets}: For each
joint separately, a single 1x1 convolution filter is learned on the
respective heatmaps from all three branches. Each filter computes
a weighted average to combine the prediction heatmaps for one specific
joint. It consists of only three weights, one for each branch, and aggregates the
different predictions based on the past, present and future.

\section{Evaluation}
\label{sec:evaluation}
We now evaluate the baseline architecture as well as the proposed additions for swimming style information 
and temporal refinement on the swimming channel data.

\paragraph{Data partitioning}
The swimming channel data consists of $24$ videos with $200$ to $400$ frames each,
leading to a total of $7146$ annotated video frames.
We partition the data into training and test sets.
Due to the cyclic nature of swimming motion, similar
poses with similar visual
appearance occur multiple times in one video. Hence, the partitioning is performed on video boundaries
such that frames from one video are either all in the training set
or in the test set. One video per swimming style is held out as the test set, while the
remaining videos are used for training. Training and test set sizes
are depicted in Table \ref{tab:swimmer_data_set_size}.

\begin{table}[t]
\noindent \begin{centering}
\begin{tabular}{l c c c c}
\hline
 & Backstr. & Breaststr. & Butterfly & Freestyle\tabularnewline
\hline \hline
Training & $1765$ & $1464$ & $1600$ & $1200$\tabularnewline

Test & $400$ & $317$ & $200$ & $200$\tabularnewline
\hline 
\end{tabular}
\par\end{centering}

\caption[Composition of the swimmer dataset.]{
  \label{tab:swimmer_data_set_size}Number of video frames
  in the training and test set.}

\end{table}

\begin{figure*}[t]
\noindent \begin{centering}
\subfloat[\label{fig:baseline_cpm_swimmer_pck}]{\noindent \begin{centering}
\includegraphics[scale=0.43]{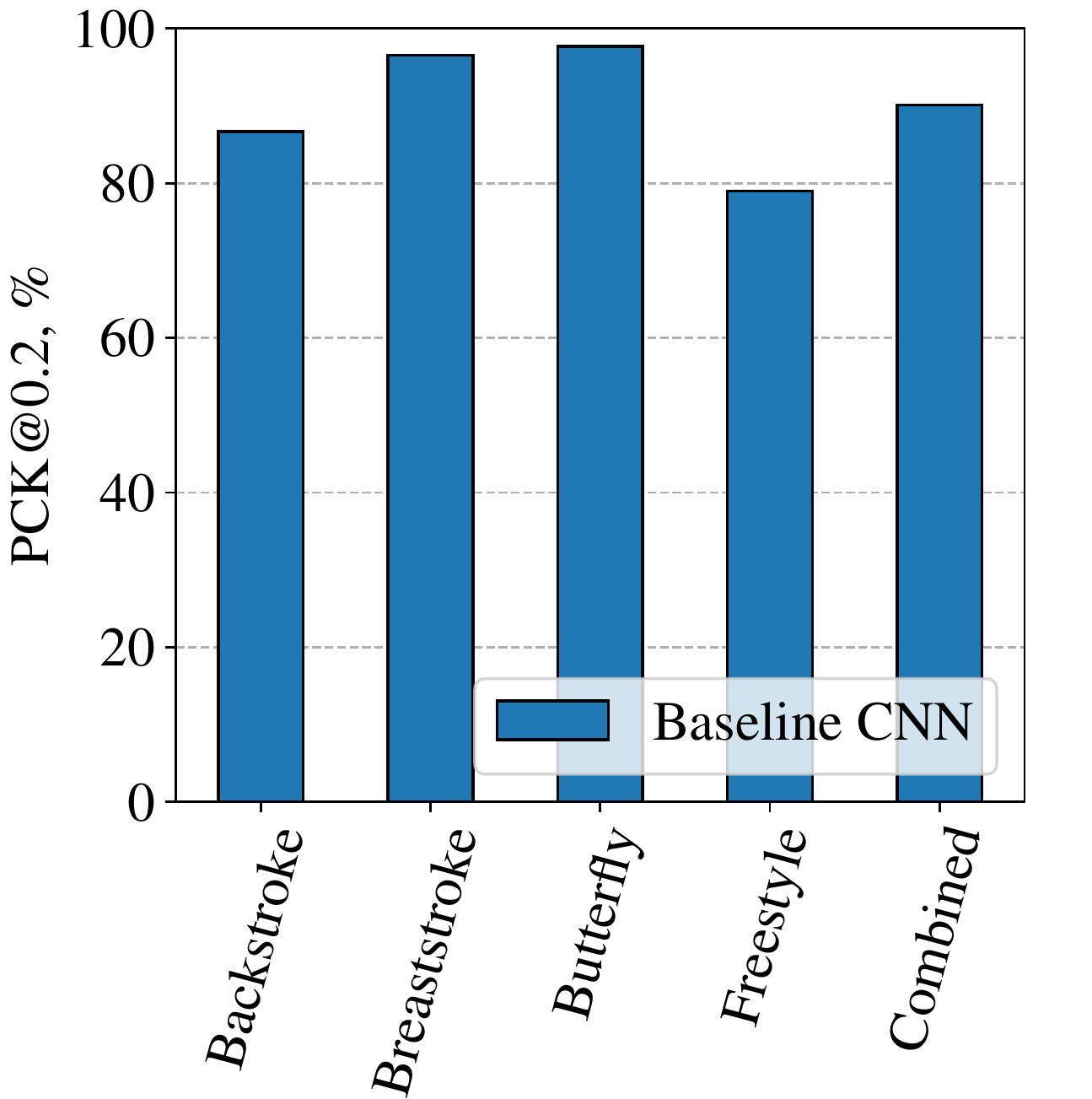}
\par\end{centering}

}\subfloat[\label{fig:baseline_cpm_swimmer_pck_joint_wise}]{\noindent \begin{centering}
\includegraphics[scale=0.43]{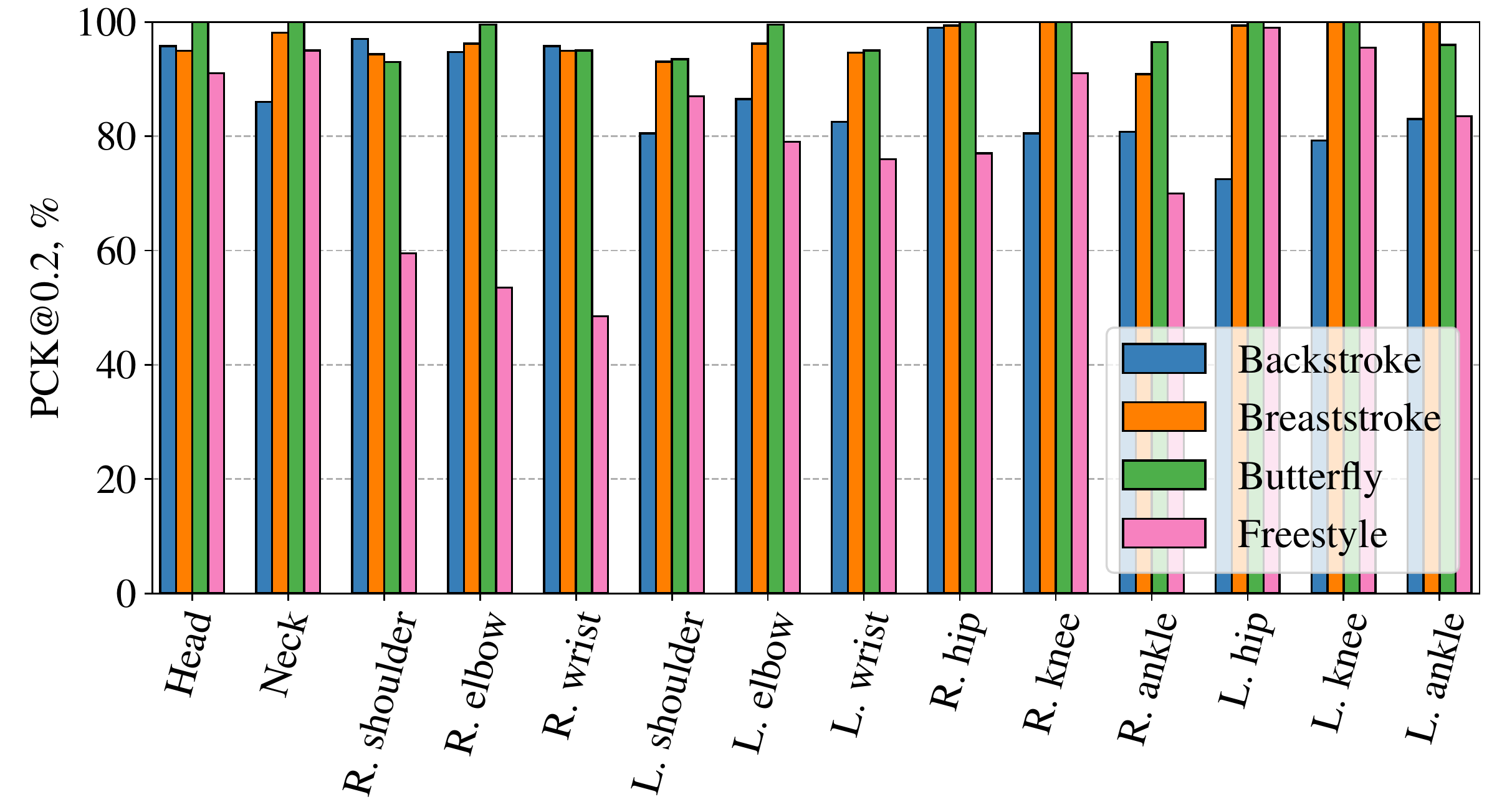}
\par\end{centering}

}
\par\end{centering}

\caption[Quantitative results of the baseline CNN on swimmer data.]{
  Results of the baseline CNN on the swimmer test set.
  PCK@0.2, (a) combined across all joints, (b) for individual joints.}
\end{figure*}

\paragraph{Metric}
For a quantitative evaluation of pose estimates we apply the \emph{Percentage
of Correct Keypoints} (PCK) metric \cite{Sapp2013_MODEC}. The PCK metric counts a joint
as correctly localized if the euclidean distance to the ground truth
location does not exceed a fixed fraction $\alpha$ of the torso diameter,
which is defined as the distance between the left side
of the hip and the right shoulder. We use $\alpha=0.2$ to compare experimental results.

\subsection{Baseline results}
\label{sec:baseline_results}

We use a 3-stage version of the CNN architecture from \cite{Wei2016_CPM} as
a performance baseline.
The network is initialized by training on the LSP dataset
for general human pose estimation in sports.
Afterwards, the network is fine-tuned on individual frames of the swimming channel data.
All experiments are performed with Caffe \cite{Jia2014_caffe} on a
NVIDIA Titan Xp GPU.

The swimmer test set performance of the baseline approach is depicted in Figure~\ref{fig:baseline_cpm_swimmer_pck}.
We achieve a detection rate of $90.1\%$
using the PCK@$0.2$ metric across the complete swimmer test set.
Table~\ref{tab:swimmer_data_set_size} reveals that the four swimming styles are not represented equally
in the test set. It contains twice as many examples for back- and
breaststroke compared to freestyle. Hence, the PCK score on the complete
test set is biased towards backstroke and breaststroke performance.
We additionally report performance on each style separately in Figure~\ref{fig:baseline_cpm_swimmer_pck}. The results on breaststroke
and butterfly are excellent with $96.6\%$ and $97.7\%$, respectively.
For backstroke and freestyle, only $86.7\%$ and $79.0\%$ of joints
are detected correctly. This indicates that symmetrical and anti-symmetrical
swimming styles pose challenges of varying difficulty. Evidently,
pose estimation for breaststroke and butterfly examples seems to be
an easier task, since left and right joints share the same annotations.

Figure~\ref{fig:baseline_cpm_swimmer_pck_joint_wise}
depicts the joint-wise PCK for each swimming style. Reliable localization
of head and neck is independent of the swimming style. Performance
on breaststroke and butterfly is good across all joints ($>90\%$).
Hardly any difference between left and right joints can be observed
here. Obviously, the CNN has learned to treat those two styles appropriately
(no left-right differentiation). For backstroke, most errors occur
on joints of the left arm and leg which are frequently occluded.
This also affects the performance on the right ankle and wrist due
to left-right confusion. Similar observations can be made for freestyle,
where occlusion of the right parts of the body is frequent. Performance
on the right arm is especially low, with a right wrist detection rate
of $48.5\%$. Figure~\ref{fig:baseline_cpm_swimmer_qualitative} shows some qualitative examples on the different swimming styles.

\subsection{Effect of swimming style labels}
\label{sec:effect_swimming_style}

\begin{figure}[t]
  \noindent
  \centering{}
  \includegraphics[scale=0.43]{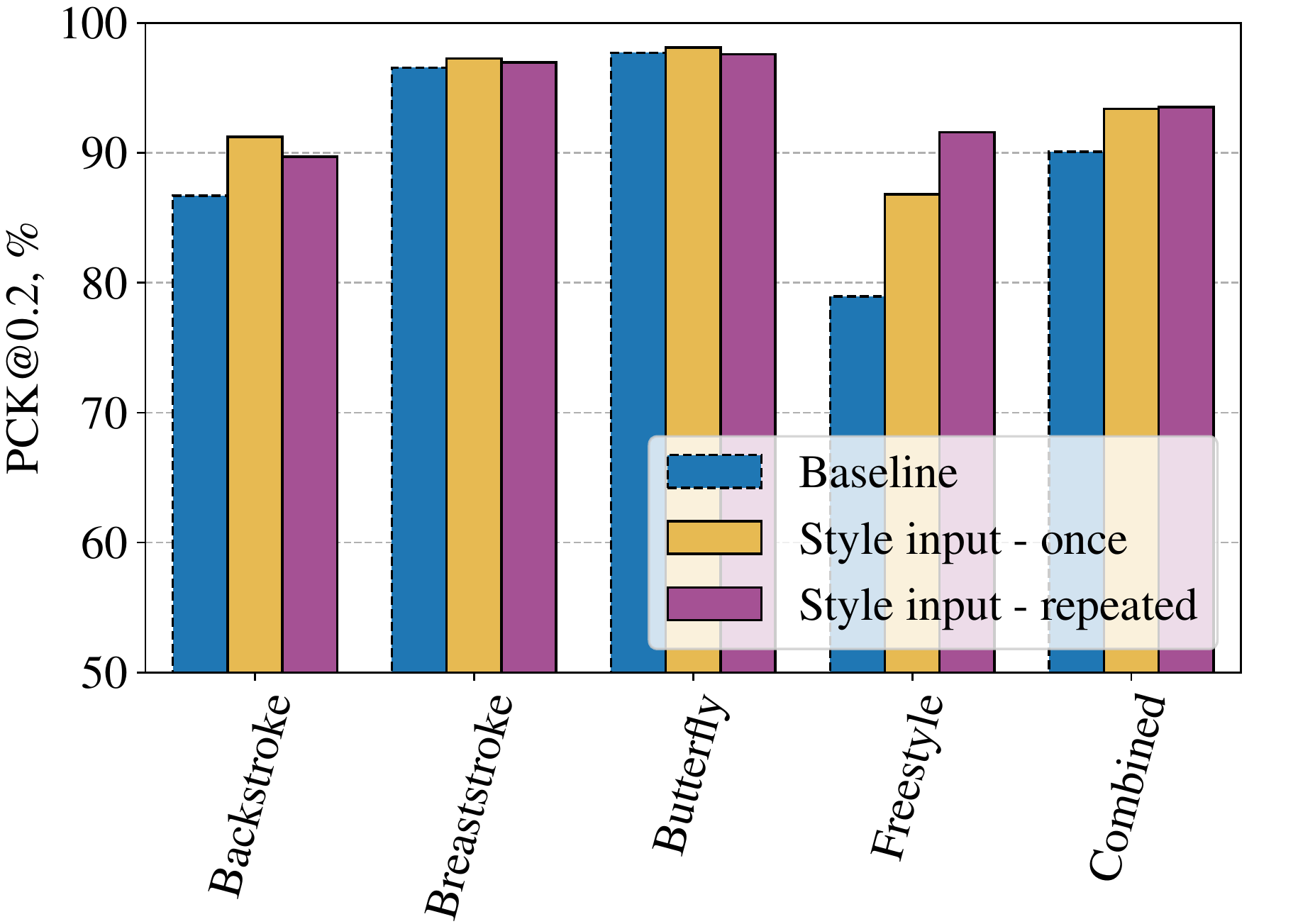}
  \caption[Result of the two CPMs with swimming style input on the test data set.]{
    \label{fig:class_input_pck}Test set results of the network variants with swimming style input.
    For convenience, baseline results are shown again (stroked).}
\end{figure}

To study the benefit of swimming style information we train the network variants with
swimming style labels from Section~\ref{sec:one_hot} in the same manner
as the baseline architecture with LSP-based initialization and fine-tuning on
the swimmer data.
New filter weights due to the added class label maps are initialized randomly.
Figure~\ref{fig:class_input_pck} depicts the test set results.
Both network variants achieve equal or better results compared to the baseline approach.
We observe no significant change for backstroke and butterfly.
This confirms that the baseline CNN can already recognize and handle
these instances appropriately.
For the anti-symmetrical strokes, the explicit swimming style information leads
to a notable improvement.
Repeated addition of the swimming style labels to subsequent convolution layers
proves to be the superior approach for freestyle, leading to a gain of $+12.6$ PCK.
The ``Style input - once '' variant however slightly dominates for the other strokes.
Detailed results are listed in Table \ref{tab:class_input_pck_diff}.
Both variants reach close to equal performance on the combined test set.
The results verify that the swimming style
(or information about a persons movement or activity in general)
is a beneficial supplementary information that can improve human pose estimation
performance significantly.
Our proposed encoding of class labels with spatially replicated
one-hot class label maps is confirmed as a viable approach for categorial inputs in CNNs.

\subsection{Pose refinement over time}
Finally, we evaluate our temporal pose refinement network from Section \ref{sec:temp_refinement_net}.
We want to analyze the effect of the refinement mechanism alone
and thus use no swimming style information for now.
The network is therefore trained and evaluated using single-frame pose estimates
obtained by the baseline CNN.
We split the training up in two phases:
First, each network branch is trained separately to predict the pose for the current frame.
The final temporal pooling layer is ignored.
In the second phase, this last layer is trained exclusively.
We use this training scheme to force each branch to predict the same target.

\begin{figure}[t]
  \noindent
  \centering{}
  \includegraphics[scale=0.43]{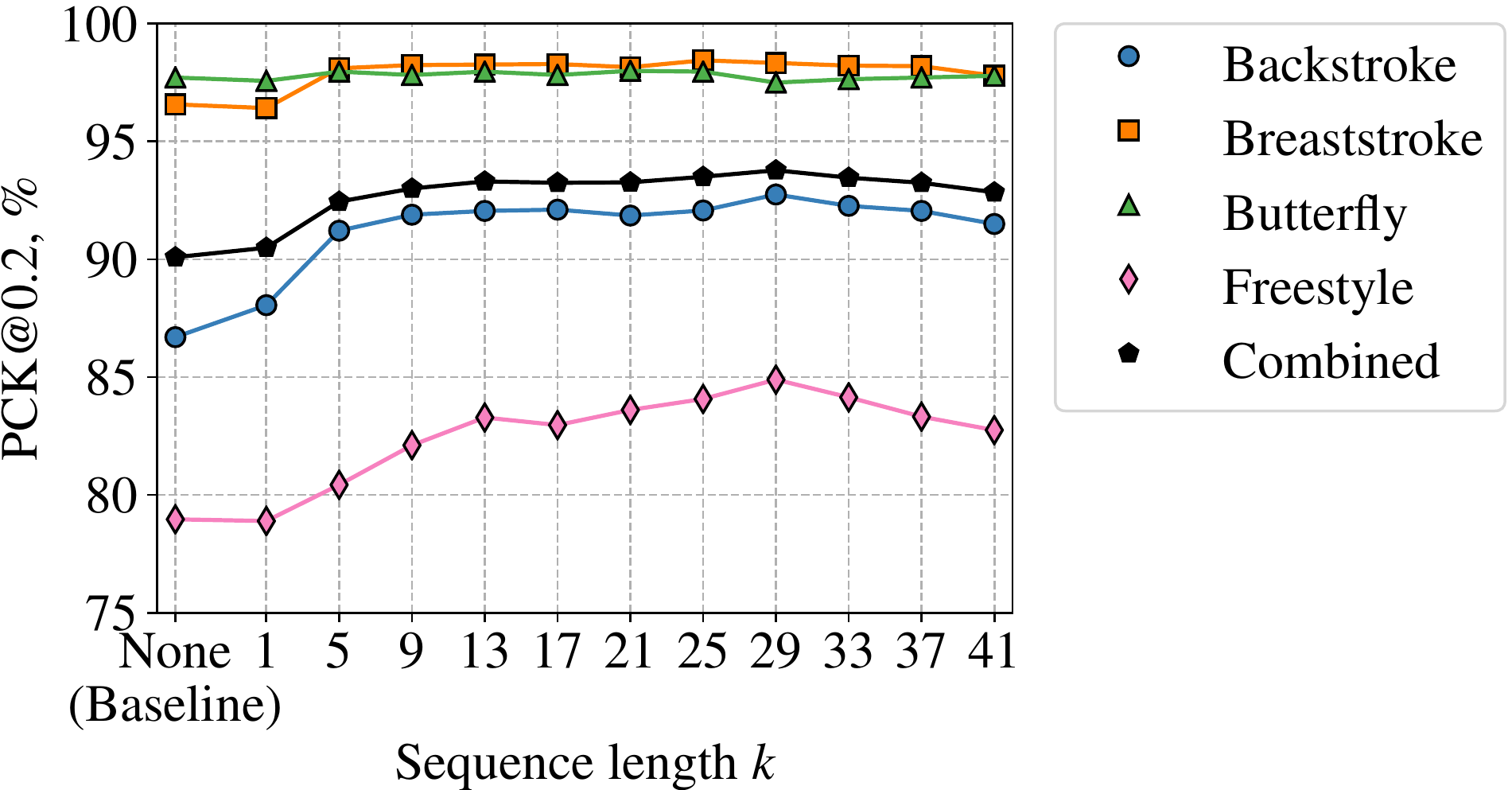}
  \caption[Results of the sequential pose refinement network, trained with data
  augmentation.]{
    \label{fig:sequence_pck}Effect of the sequence length $k$ on the temporal refinement network, evaluated on the swimmer test set. 
    Input sequences are of length $1$ to $41$, but include pose estimates from every second frame only.}
\end{figure}

We expect that the sequence length parameter $k$ has a direct influence
on pose estimation (refinement) performance. We evaluate multiple values for $k \in [1,41]$ and show the results in Figure~\ref{fig:sequence_pck}.
The case $k=1$ can be seen as a verification experiment. With no sequential
information at all, the outer network branches can be ignored and we only have a fourth network stage in
the baseline CNN architecture. \cite{Wei2016_CPM} argue that additional
stages up to a total of $s=6$ are beneficial, at least when evaluated
on the LSP dataset.
The results with $k=1$ reach and slightly surpass
the baseline performance. They confirm that an additional and separately
trained stage even without any past or future pose input is a viable option.
When the network does use sequential pose estimates, \ie $k>1$, performance increases together with the sequence length. 
For $k=29$, we achieve $93.8\%$ on the combined test set, an additional $+1.9$ PCK compared to the baseline result.
This improvement is almost entirely gained at
backstroke and freestyle, with $+6.0$ PCK each.
With even longer input sequences, performance starts to decline again for some of the swimming styles. 
Additionally, we obseve that the increasingly larger network input renders the training less stable.

With the simple design of the temporal pooling layer at the end of the refinement network, its weights
can be analyzed to see what influence each network branch has on the
final output. The layer consists of $3$ weights (and one bias term)
for each joint, and computes a weighted average on the
past, present and future prediction heatmaps. Table \ref{tab:temporal_pooling_weights}
depicts the layer weights averaged over all joints. Similar
importance is assigned to all three branches. 
Consequently, the predictions based on past and future information (\ie based on $\mathbf{z}_t$) 
are vital for the quality of the final network prediction -- a design goal of this architecture.

\begin{table}[t]
\noindent \centering{}%
\begin{tabular}{l c c c}
\hline 
& \texttt{p\_conv6} & \texttt{conv9} & \texttt{f\_conv6}\tabularnewline
\hline 
\hline 
Avg. weight & $0.289$ & $0.300$ & $0.306$\tabularnewline
\hline 
\end{tabular}\caption[Learned weights in the temporal pooling layer.]{
  \label{tab:temporal_pooling_weights}Temporal pooling calculates
  a joint-wise weighted average on the output layers of the three network
  branches. The table shows the weights averaged over all joints.}
\end{table}

\begin{table*}[t]
\noindent \centering{}%
\begin{tabular}{l c c c c c}
\hline
 & Backstroke & Breaststroke & Butterfly & Freestyle & Combined\tabularnewline
\hline \hline
Baseline CNN \cite{Wei2016_CPM} & $86.7$ & $96.6$ & $97.7$ & $79.0$ & $90.1$\tabularnewline
\hline 
\makecell[cl]{Style input - once} & $91.2$ & $97.3$ & $\mathbf{98.1}$ & $86.8$ & $93.4$\tabularnewline
\hline 
\makecell[cl]{Style input - repeated} & $89.7$ & $97.0$ & $97.6$ & $91.6$ & $93.5$\tabularnewline
\hline 
\makecell[cl]{Temporal pose refinement, $k=29$} & $\mathbf{92.7}$ & $98.3$ & $97.5$ & $85.0$ & $93.8$\tabularnewline
\hline 
\makecell[cl]{Temporal pose refinement, $k=29$, +swimming style} & $92.4$ & $\mathbf{98.5}$ & $97.7$ & $\mathbf{95.7}$ & $\mathbf{95.7}$\tabularnewline
\hline 
\end{tabular}\caption[Swimming style input CPM results compared to single-style CPMs.]{
  \label{tab:class_input_pck_diff} Summary of the PCK@0.2 scores on the swimmer test set.}
\end{table*}

\begin{figure*}[t]
  \noindent
  \centering{}
  \includegraphics[width=0.97\textwidth, keepaspectratio]{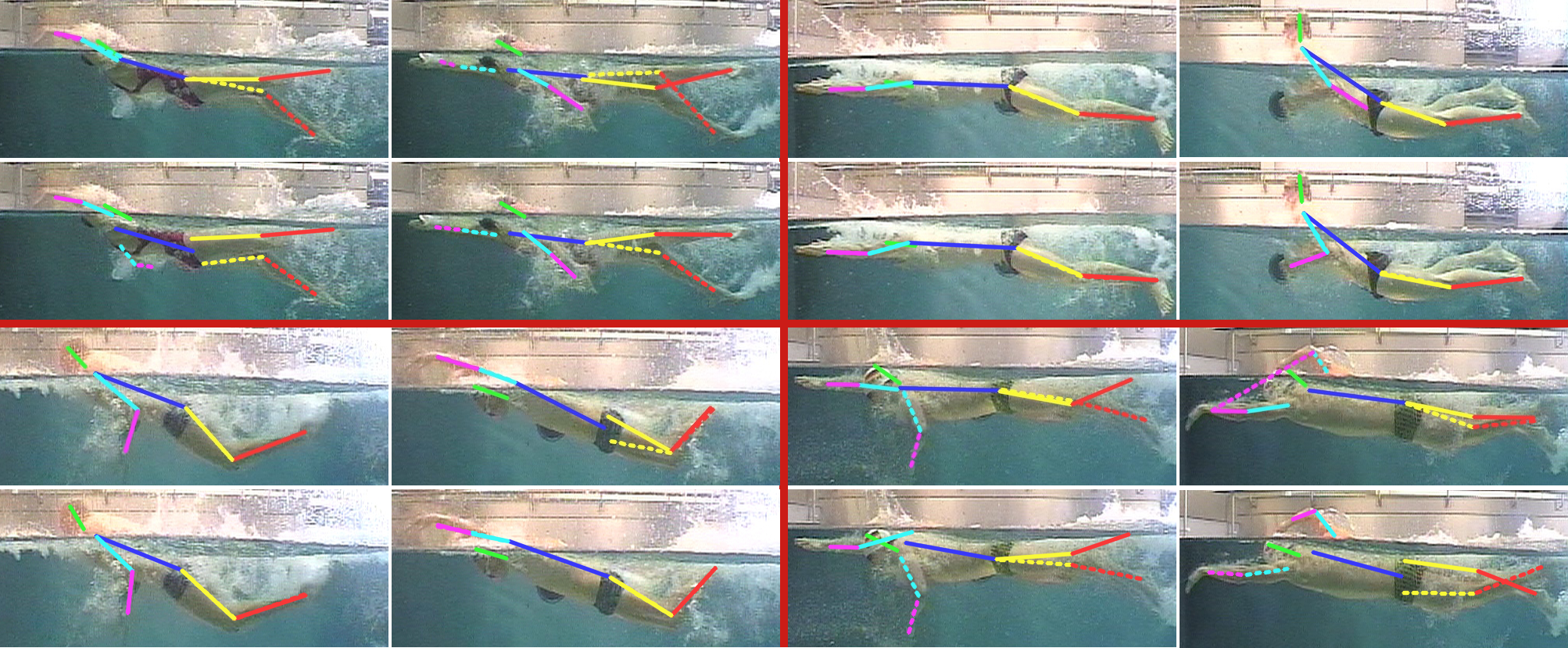}
  \caption[Qualitative results of baseline CPM on swimmer data.]{
    \label{fig:baseline_cpm_swimmer_qualitative}Qualitative results
    on the swimming channel data. The pose is visualized by drawing connections between respective joints.
    The parts on the left side of the body are stroked. Each quadrant contains examples
    from one swimming style: backstroke, breaststroke, freestyle, butterfly, from top left in clockwise order.
    Odd rows depict results from the baseline CNN, 
    even rows depict the results on the same instances using our best architecture with temporal pose refinement and swimming style information.}
\end{figure*}

\begin{figure}[t]
  \noindent
  \centering{}
  \includegraphics[width=0.4825\textwidth]{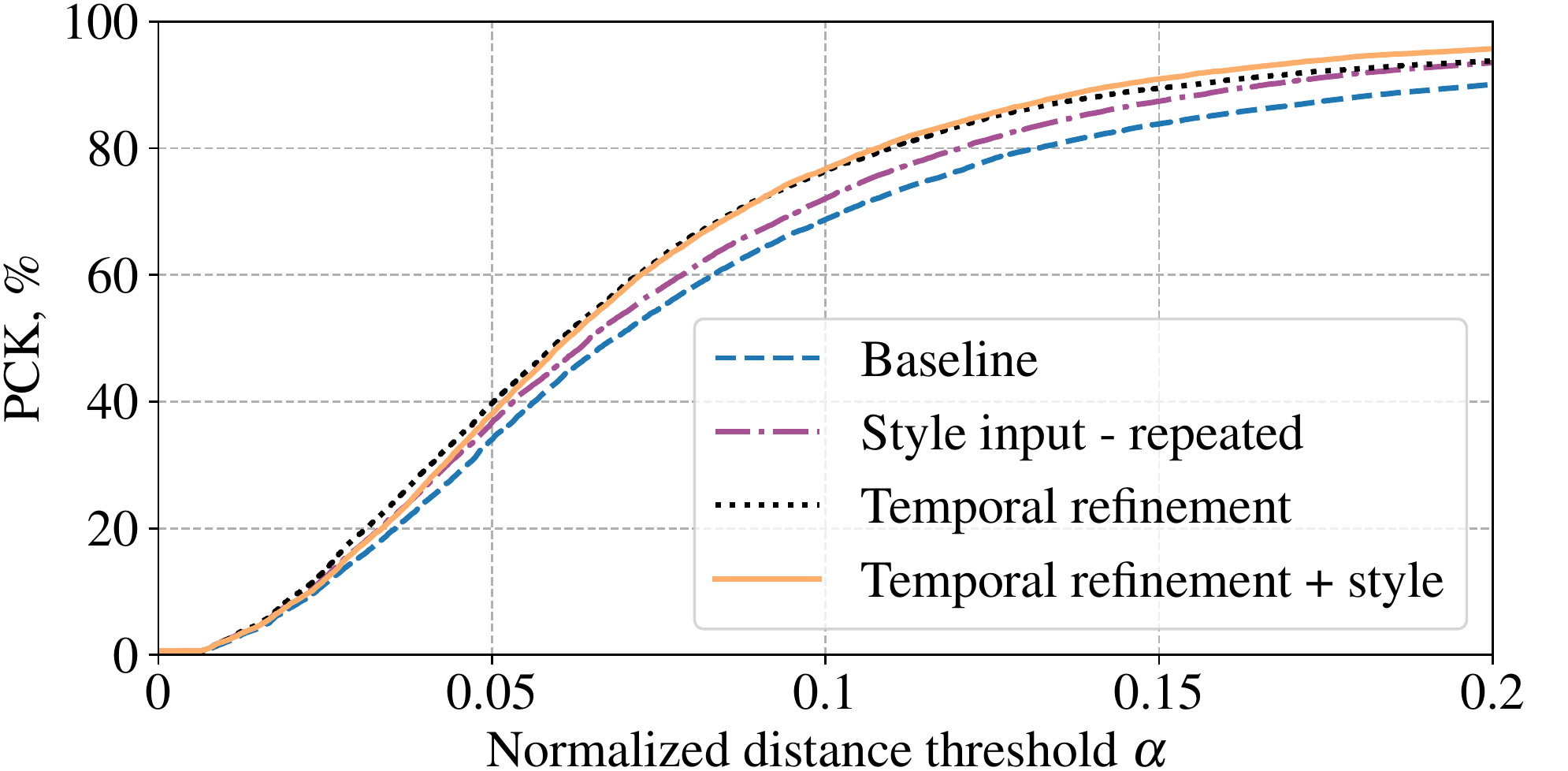}
  \caption[Result on the PCK range.]{
    \label{fig:pck_range}Results of all major network variants with a PCK threshold range $\alpha \in [0, 0.2]$}
\end{figure}

\subsection{Combined architecture}

It is straightforward to combine the findings of the preceding evaluation
into a single architecture with swimming style information and temporal refinement.
We simply combine the ``Style input - repeated'' architecture
from Section~\ref{sec:one_hot} for single-frame pose estimates and the
temporal refinement network (identically augmented with swimming style information).
The test set result is listed in Table~\ref{tab:class_input_pck_diff}.
In this configuration we can further boost the pose estimation performance to
excellent $95.7\%$ on the complete test set.
It shows that the swimming style and the joint predictions over time are both very
relevant and complementary sources of information.
Figure~\ref{fig:baseline_cpm_swimmer_qualitative} depicts a comparison to the baseline results on some qualitative examples 
with different swimming styles. We additionally provide the results of our major network variants with varying PCK distance thresholds in Figure~\ref{fig:pck_range}. It shows that our approaches can equally improve performance for higher precision joint detection.

\section{Conclusion}
\label{sec:conclusion}

This paper presents two extensions to the standard Convolutional Pose Machine model to deal with the challenges of human pose estimation for swimmers in the water.
The first extension is based on the reduction of the subspace of possible pose configurations with swimmming style information.
We formulate a spatially redundant one-hot encoding of class labels that allows the network to learn swimming style specific filters.
This principle can be applied to any from of activity information.
The second extension focuses on the temporal aspect of swimmer videos.
We present a two-step approach where initial pose estimates are refined in fixed-length sequences by a separate CNN module.
The results show that the network benefits from long sequences, indicating its ability to predict and refine poses forward and backward in time.
Both extensions can be combined easily. We show that we clearly improve over the baseline CPM architecture.
Our findings are directly applicable to other sports where reliable pose esimates are essential for effective performance analysis.

\section*{Acknowledgements}
\label{sec:acknowledgements}
This work was funded by the Federal Institute for Sports Science based on a resolution of the German Bundestag. We would like to thank the Institute for Applied Training Science (IAT) Leipzig for providing the video data.

{\small
\bibliographystyle{ieee}
\bibliography{references}

\begin{thebibliography}{10}\itemsep=-1pt

\bibitem{Andriluka2014_MPII}
M.~Andriluka, L.~Pishchulin, P.~Gehler, and B.~Schiele.
\newblock {2D} human pose estimation: New benchmark and state of the art
  analysis.
\newblock In {\em The IEEE Conference on Computer Vision and Pattern
  Recognition (CVPR)}, June 2014.

\bibitem{andriluka2009_PictStructs_Revisited}
M.~Andriluka, S.~Roth, and B.~Schiele.
\newblock Pictorial structures revisited: People detection and articulated pose
  estimation.
\newblock In {\em The IEEE Conference on Computer Vision and Pattern
  Recognition (CVPR)}, pages 1014--1021, June 2009.

\bibitem{Belagiannis2016_Recurrent_HPE}
V.~Belagiannis and A.~Zisserman.
\newblock Recurrent human pose estimation.
\newblock In {\em 2017 12th IEEE International Conference on Automatic Face
  Gesture Recognition (FG 2017)}, pages 468--475, May 2017.

\bibitem{Cao2016_CPM_MultiPerson}
Z.~Cao, T.~Simon, S.-E. Wei, and Y.~Sheikh.
\newblock Realtime multi-person {2D} pose estimation using part affinity
  fields.
\newblock In {\em The IEEE Conference on Computer Vision and Pattern
  Recognition (CVPR)}, July 2017.

\bibitem{Carreira2016_error_feedback}
J.~Carreira, P.~Agrawal, K.~Fragkiadaki, and J.~Malik.
\newblock Human pose estimation with iterative error feedback.
\newblock In {\em The IEEE Conference on Computer Vision and Pattern
  Recognition (CVPR)}, June 2016.

\bibitem{Fastovets2013_PE}
M.~Fastovets, J.-Y. Guillemaut, and A.~Hilton.
\newblock Athlete pose estimation from monocular tv sports footage.
\newblock In {\em The IEEE Conference on Computer Vision and Pattern
  Recognition (CVPR) Workshops}, June 2013.

\bibitem{Felzenszwalb2005_PictStructs}
P.~F. Felzenszwalb and D.~P. Huttenlocher.
\newblock Pictorial structures for object recognition.
\newblock {\em International Journal of Computer Vision}, 61(1):55--79, 2005.

\bibitem{Gade2013_SportsType}
R.~Gade and T.~B. Moeslund.
\newblock Sports type classification using signature heatmaps.
\newblock In {\em The IEEE Conference on Computer Vision and Pattern
  Recognition (CVPR) Workshops}, June 2013.

\bibitem{Gkioxari_2016}
G.~Gkioxari, A.~Toshev, and N.~Jaitly.
\newblock Chained predictions using convolutional neural networks.
\newblock In {\em European Conference on Computer Vision}, pages 728--743.
  Springer, 2016.

\bibitem{Hasegawa2016}
K.~Hasegawa and H.~Saito.
\newblock Synthesis of a stroboscopic image from a hand-held camera sequence
  for a sports analysis.
\newblock {\em Computational Visual Media}, 2(3):277--289, 2016.

\bibitem{Hwang_2017_CVPR_Workshops}
J.~Hwang, S.~Park, and N.~Kwak.
\newblock Athlete pose estimation by a global-local network.
\newblock In {\em The IEEE Conference on Computer Vision and Pattern
  Recognition (CVPR) Workshops}, July 2017.

\bibitem{Insafutdinov2016_DeeperCut}
E.~Insafutdinov, L.~Pishchulin, B.~Andres, M.~Andriluka, and B.~Schiele.
\newblock {DeeperCut}: A deeper, stronger, and faster multi-person pose
  estimation model.
\newblock In {\em European Conference on Computer Vision}, pages 34--50.
  Springer, 2016.

\bibitem{Jia2014_caffe}
Y.~Jia, E.~Shelhamer, J.~Donahue, S.~Karayev, J.~Long, R.~Girshick,
  S.~Guadarrama, and T.~Darrell.
\newblock Caffe: Convolutional architecture for fast feature embedding.
\newblock In {\em Proceedings of the 22nd ACM international conference on
  Multimedia}, pages 675--678. ACM, 2014.

\bibitem{Johnson10_LSP}
S.~Johnson and M.~Everingham.
\newblock Clustered pose and nonlinear appearance models for human pose
  estimation.
\newblock In {\em Proceedings of the British Machine Vision Conference}, 2010.

\bibitem{Moeslund2015_CV_Sports}
T.~B. Moeslund, G.~Thomas, and A.~Hilton.
\newblock {\em Computer vision in sports}.
\newblock Springer, 2015.

\bibitem{Newell2016_Hourglass}
A.~Newell, K.~Yang, and J.~Deng.
\newblock Stacked hourglass networks for human pose estimation.
\newblock In {\em European Conference on Computer Vision}, pages 483--499.
  Springer, 2016.

\bibitem{Nibali_2017_CVPR_Workshops}
A.~Nibali, Z.~He, S.~Morgan, and D.~Greenwood.
\newblock Extraction and classification of diving clips from continuous video
  footage.
\newblock In {\em The IEEE Conference on Computer Vision and Pattern
  Recognition (CVPR) Workshops}, July 2017.

\bibitem{Pfister2015_Flowing_ConvNets}
T.~Pfister, J.~Charles, and A.~Zisserman.
\newblock Flowing {Conv{--}Nets} for human pose estimation in videos.
\newblock In {\em Proceedings of the IEEE International Conference on Computer
  Vision}, pages 1913--1921, 2015.

\bibitem{Pishchulin2016_DeepCut}
L.~Pishchulin, E.~Insafutdinov, S.~Tang, B.~Andres, M.~Andriluka, P.~V. Gehler,
  and B.~Schiele.
\newblock {DeepCut}: Joint subset partition and labeling for multi person pose
  estimation.
\newblock In {\em The IEEE Conference on Computer Vision and Pattern
  Recognition (CVPR)}, June 2016.

\bibitem{Ramakrishna2014_PoseMachines}
V.~Ramakrishna, D.~Munoz, M.~Hebert, J.~A. Bagnell, and Y.~Sheikh.
\newblock Pose machines: Articulated pose estimation via inference machines.
\newblock In {\em European Conference on Computer Vision}, pages 33--47.
  Springer, 2014.

\bibitem{Sapp2013_MODEC}
B.~Sapp and B.~Taskar.
\newblock Modec: Multimodal decomposable models for human pose estimation.
\newblock In {\em Proceedings of the IEEE Conference on Computer Vision and
  Pattern Recognition}, pages 3674--3681, 2013.

\bibitem{Sha2013_swimmer_loc}
L.~Sha, P.~Lucey, S.~Morgan, D.~Pease, and S.~Sridharan.
\newblock Swimmer localization from a moving camera.
\newblock In {\em 2013 International Conference on Digital Image Computing:
  Techniques and Applications (DICTA)}, pages 1--8, Nov 2013.

\bibitem{Song_2017_CVPR}
J.~Song, L.~Wang, L.~Van~Gool, and O.~Hilliges.
\newblock Thin-slicing network: A deep structured model for pose estimation in
  videos.
\newblock In {\em The IEEE Conference on Computer Vision and Pattern
  Recognition (CVPR)}, July 2017.

\bibitem{Soomro_2015_Tracking}
K.~Soomro, S.~Khokhar, and M.~Shah.
\newblock Tracking when the camera looks away.
\newblock In {\em The IEEE International Conference on Computer Vision (ICCV)
  Workshops}, December 2015.

\bibitem{Tatarchenko2016_Multi_View}
M.~Tatarchenko, A.~Dosovitskiy, and T.~Brox.
\newblock Multi-view {3D} models from single images with a convolutional
  network.
\newblock In {\em European Conference on Computer Vision}, pages 322--337.
  Springer, 2016.

\bibitem{Toshev2014_DeepPose}
A.~Toshev and C.~Szegedy.
\newblock {DeepPose}: Human pose estimation via deep neural networks.
\newblock In {\em Proceedings of the IEEE Conference on Computer Vision and
  Pattern Recognition}, pages 1653--1660, 2014.

\bibitem{Turchini2015_Sport_Activities}
F.~Turchini, L.~Seidenari, and A.~Del~Bimbo.
\newblock Understanding sport activities from correspondences of clustered
  trajectories.
\newblock In {\em The IEEE International Conference on Computer Vision (ICCV)
  Workshops}, December 2015.

\bibitem{Victor_2017_CVPR_Workshops}
B.~Victor, Z.~He, S.~Morgan, and D.~Miniutti.
\newblock Continuous video to simple signals for swimming stroke detection with
  convolutional neural networks.
\newblock In {\em The IEEE Conference on Computer Vision and Pattern
  Recognition (CVPR) Workshops}, July 2017.

\bibitem{Wang_2013_CVPR}
F.~Wang and Y.~Li.
\newblock Beyond physical connections: Tree models in human pose estimation.
\newblock In {\em The IEEE Conference on Computer Vision and Pattern
  Recognition (CVPR)}, June 2013.

\bibitem{Wei2016_CPM}
S.-E. Wei, V.~Ramakrishna, T.~Kanade, and Y.~Sheikh.
\newblock Convolutional pose machines.
\newblock In {\em Proceedings of the IEEE Conference on Computer Vision and
  Pattern Recognition}, pages 4724--4732, 2016.

\bibitem{Wei2015_Ball_Ownership}
X.~Wei, L.~Sha, P.~Lucey, P.~Carr, S.~Sridharan, and I.~Matthews.
\newblock Predicting ball ownership in basketball from a monocular view using
  only player trajectories.
\newblock In {\em The IEEE International Conference on Computer Vision (ICCV)
  Workshops}, December 2015.

\bibitem{Yang_2016_CVPR}
W.~Yang, W.~Ouyang, H.~Li, and X.~Wang.
\newblock End-to-end learning of deformable mixture of parts and deep
  convolutional neural networks for human pose estimation.
\newblock In {\em The IEEE Conference on Computer Vision and Pattern
  Recognition (CVPR)}, June 2016.

\bibitem{Zecha2015_KeyPose}
D.~Zecha and R.~Lienhart.
\newblock Key-pose prediction in cyclic human motion.
\newblock In {\em Applications of Computer Vision (WACV), 2015 IEEE Winter
  Conference on}, pages 86--93. IEEE, 2015.

\bibitem{Zhang_2015_ICCV}
D.~Zhang and M.~Shah.
\newblock Human pose estimation in videos.
\newblock In {\em The IEEE International Conference on Computer Vision (ICCV)},
  December 2015.

\end{thebibliography}
}

\end{document}